\documentclass[runningheads]{llncs}
\usepackage{graphicx}
\usepackage{amsmath,amsfonts,amssymb}
\usepackage{algorithm,algpseudocode}
\usepackage{array}
\usepackage[caption=false,font=normalsize,labelfont=sf,textfont=sf]{subfig}
\usepackage{textcomp}
\usepackage{stfloats}
\usepackage{url}
\usepackage{verbatim}
\usepackage{graphicx}
\usepackage{balance}
\usepackage{afterpage}
\usepackage{tabularx}
\usepackage[noadjust]{cite}
\usepackage[colorlinks,bookmarksopen,bookmarksnumbered,citecolor=blue,urlcolor=blue]{hyperref}
\usepackage{booktabs}
\usepackage{makecell}
\usepackage{multirow}

\begin{document}
\title{Multimodal Dataset for Localization, Mapping and Crop Monitoring in Citrus Tree Farms}

%
\author{Hanzhe Teng \and Yipeng Wang \and Xiaoao Song \and Konstantinos Karydis}
%
\authorrunning{H. Teng et al.}
\titlerunning{Multimodal Dataset in Citrus Tree Farms}
%
\institute{University of California Riverside, Riverside, CA 92521, USA \\
\email{\{hteng007,ywang1040,xsong036,karydis\}@ucr.edu}
}

\maketitle              
%
%
\begin{abstract}
In this work we introduce the CitrusFarm dataset, a comprehensive multimodal sensory dataset collected by a wheeled mobile robot operating in agricultural fields.
The dataset offers stereo RGB images with depth information, as well as monochrome, near-infrared and thermal images, presenting diverse spectral responses crucial for agricultural research.
Furthermore, it provides a range of navigational sensor data encompassing wheel odometry, LiDAR, inertial measurement unit (IMU), and GNSS with Real-Time Kinematic (RTK) as the centimeter-level positioning ground truth.
The dataset comprises seven sequences collected in three fields of citrus trees, featuring various tree species at different growth stages, distinctive planting patterns, as well as varying daylight conditions. 
It spans a total operation time of $1.7$\;hours, covers a distance of $7.5$\;km, and constitutes $1.3$\;TB of data. 
We anticipate that this dataset can facilitate the development of autonomous robot systems operating in agricultural tree environments, especially for localization, mapping and crop monitoring tasks.
Moreover, the rich sensing modalities offered in this dataset can also support research in a range of robotics and computer vision tasks, such as place recognition, scene understanding, object detection and segmentation, and multimodal learning. 
The dataset, in conjunction with related tools and resources, is made publicly available at \url{https://github.com/UCR-Robotics/Citrus-Farm-Dataset}.

\end{abstract}

\section{Introduction}
Crop monitoring is an essential and critical component in precision agriculture. Its value is that it can provide growers and agronomists with different pieces of information that are useful for determining a range of indicators, such as plant growth~\cite{kim2021stereo}, water stress level and health condition~\cite{zhao2015ndvi}, to name a few. This can lead to more informed future decisions on irrigation, disease prevention, pest control, fruit harvesting and ultimately higher yields~\cite{vougioukas2019agricultural}.
Given the dynamic nature of changes that happen in the field, crop monitoring tasks need to be performed regularly, a process that can quickly become time- and labor-intensive and scale poorly both spatially and temporally~\cite{bechar2016agricultural}. 
These challenges can be addressed to a certain extent by automating parts of the overall process with mobile robots. 

Datasets are an essential tool to help develop autonomous robot systems in agricultural environments.
Navigational sensor data is crucial for the development of accurate localization and mapping algorithms, which can enable a range of fully automated tasks such as soil conductivity measurement~\cite{merrick2021case} and physical sample retrieval~\cite{merrick2022iros}.
On the other hand, multispectral images are particularly valuable in agricultural research, as they can provide informative measurements for computing domain-specific indices. For example, the Normalized Difference Vegetation Index (NDVI) can be computed from red and near-infrared spectral channels and serves as an indicator of plant health~\cite{zhao2015ndvi}.
If both sensing modalities are available, a detailed and accurate map with multimodal data (e.g., thermal, NDVI) can be constructed by an automated mobile robot for crop monitoring on a regular basis.
However, there currently exist limited public datasets that provide \emph{both multispectral images and navigational sensor data} to facilitate the development of such autonomous robot systems.

To this end, we present the CitrusFarm dataset, a comprehensive multi-sensor dataset in the citrus tree farms.
The dataset comprises seven sequences collected in three fields of citrus trees, featuring various tree species at different growth stages, distinctive planting patterns, as well as varying daylight conditions. 
The dataset offers stereo RGB images with depth information, as well as monochrome, near-infrared and thermal images, presenting diverse spectral responses crucial for agricultural research.
Furthermore, it provides a range of navigational sensor data encompassing wheel odometry, LiDAR, inertial measurement unit (IMU), and GNSS with Real-Time Kinematic (RTK) as the centimeter-level positioning ground truth.
In comparison to related works, our CitrusFarm dataset provides a total of nine sensing modalities, thus enabling significant potential for multimodal and crossmodal research in the robotics and computer vision community.

We anticipate that this dataset can facilitate the development of autonomous robot systems operating in agricultural tree environments, primarily for localization, mapping and navigation tasks.
With both multispectral and navigational data, automated multimodal mapping becomes possible for crop monitoring on a regular basis. Given the rich sensing modalities, sensor fusion can be conducted with any combination of sensing sources of interest. This can advance the research on localization in challenging unstructured agricultural environments, and enable a full range of automated agricultural operations contingent on accurate localization. 
Lastly, this dataset can serve as the training and testing data in machine learning tasks, and support research in a variety of robotics and computer vision applications, such as scene understanding, segmentation, object detection, place recognition and multimodal learning.

\section{Related Works}
Agricultural robotics datasets that provide both multispectral images and navigational sensor data are scarce. Lu and Young reviewed public agricultural image datasets for weed control, fruit detection and miscellaneous applications, without navigational sensors involved~\cite{lu2020survey}.
In the following, \emph{we focus on public datasets that contain both images and navigational data}. Overall, such datasets can be categorized into two broad types.

\textbf{Datasets recorded following the same route over time:}
Chebrolu et al. recorded the dataset on a sugar beet farm over a period of three months, aiming to advance research on plant classification, localization and mapping~\cite{chebrolu2017sugarbeets}.
%
Bender et al. presented a multimodal dataset that contains weekly scans of cauliflower and broccoli covering a ten-week growth cycle from transplant to harvest using the Ladybird agricultural robot~\cite{bender2020Ladybird}. In addition, that dataset also provides physical characteristics of the crop, soil sensor data and environmental data from a weather station.
%
Polvara et al. collected multi-sensor data following the same route in a vineyard environment over a time span of six months; the resulting dataset can be used for long-term localization and mapping, phenotyping and crop mapping tasks~\cite{blt2022dataset}.

\textbf{Datasets collected in a variety of scenes:}
Pire et al. collected six sequences of multi-sensor data on soybean fields, with the focus being on visual localization and mapping under various challenging situations~\cite{pire2019rosario}.
%
G{\"u}ldenring et al. focused on the problem of weed control using vision-equipped robots and presented the RumexWeeds dataset that comprises color images with annotations and navigational sensing data associated with each image frame~\cite{guldenring2023rumexweeds}.
%
Ali et al. recorded the dataset in a forest environment in various seasons and daylight conditions using a sensor rig mounted on a car vehicle~\cite{ali2020finnforest}.

Our work fits into the second category, and aims to diversify the data collection process in the citrus tree environments. 
We collected seven sequences in three agricultural fields featuring various tree species at different growth stages, distinctive planting patterns, as well as varying daylight conditions, all while following various routes in the fields. 
More details about our data collection process are to be discussed in Section~\ref{sec_data_collection}.

Moreover, in comparison to related works, our CitrusFarm dataset provides a total of nine sensing modalities, thereby unlocking vast potential for multimodal research in robotics and computer vision.
Among the nine modalities, the presence of LiDAR data distinguishes this work from the offering geo-referenced images alone, and multispectral images further provide with a unique perceptual perspective.
A detailed overview of the sensing modalities encompassed within each dataset is presented in Table~\ref{table_dataset_related_works}.

\begin{table}[t]
\caption{Summary of Available Sensing Modalities in Related Public Datasets and Comparison with Our Developed Dataset.}
\label{table_dataset_related_works}
\vspace{-20pt}
\begin{center}
\resizebox{\columnwidth}{!}{
\begin{tabular}{cccccccccc}
\toprule
Dataset & Mono & RGB & Depth & NIR & Thermal & LiDAR & Wheel-Odom & IMU & GPS(-RTK) \\
\midrule
Chebrolu et al.~\cite{chebrolu2017sugarbeets}  &   & \checkmark & \checkmark & \checkmark &   & \checkmark & \checkmark &   & \checkmark  \\
Bender et al.~\cite{bender2020Ladybird} &  \checkmark  & \checkmark  &    & \checkmark & \checkmark &  &  &  \checkmark & \checkmark \\
Polvara et al.~\cite{blt2022dataset} &  & \checkmark  &  \checkmark  &  &  & \checkmark & \checkmark & \checkmark & \checkmark \\
Pire et al.~\cite{pire2019rosario} &  & \checkmark &  &  &  &  & \checkmark & \checkmark & \checkmark \\
G{\"u}ldenring et al.~\cite{guldenring2023rumexweeds} &   & \checkmark &  &  &   &   & \checkmark & \checkmark & \checkmark \\
Ali et al.~\cite{ali2020finnforest} &  & \checkmark &  &  &  &  &  &  \checkmark & \checkmark \\
Ours & \checkmark & \checkmark & \checkmark & \checkmark & \checkmark & \checkmark & \checkmark & \checkmark & \checkmark \\
\bottomrule
\end{tabular}
}
\end{center}
\vspace{-8pt}
{\footnotesize
}
\end{table}

\section{Mobile Robot Setup}
We employ the Clearpath Jackal wheeled mobile robot to collect multi-sensor data in the agricultural fields.
Jackal is a differential-drive robot that follows the unicycle kinematic model, which can take linear and angular velocities as the control input. It can support in-place turns and a maximum linear speed of $2$\;m/s.
The onboard computer installed on the robot is a Jetson Xavier AGX, with $32$\;GB RAM, $2$\;TB storage and Ubuntu 18.04 operating system.
The Robot Operating System (ROS) is utilized as the middleware to connect with all sensors and record data.

\begin{figure}[!t]
\begin{center}
\includegraphics[width=0.9\linewidth]{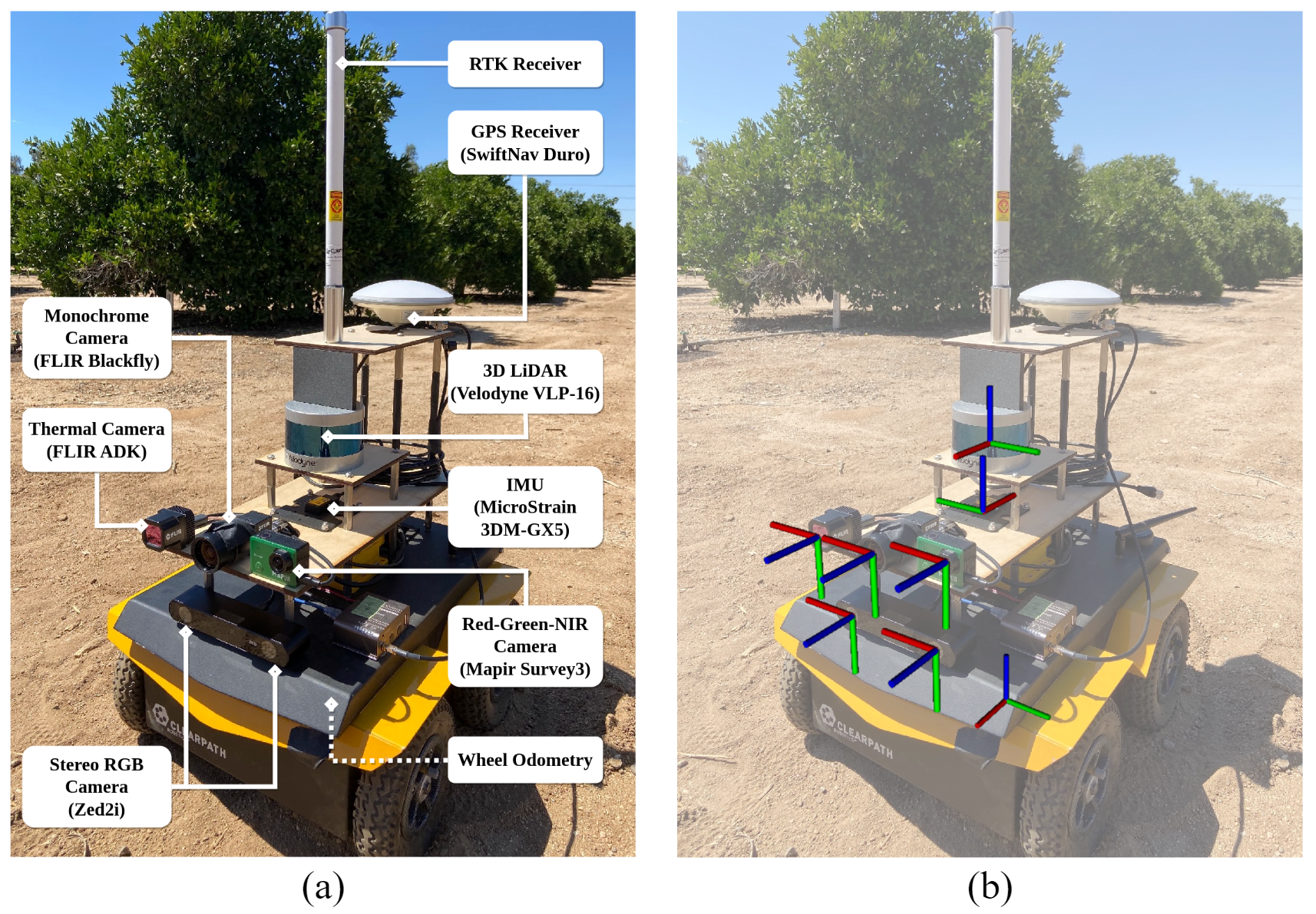}
\caption{(a) The Clearpath Jackal mobile robot with a variety of sensors mounted onboard, ready to collect data in the field. (b) The corresponding reference frames of five camera lenses, IMU, LiDAR and wheel odometry superimposed on the Jackal robot. Relative poses between frames are real parameters after extrinsic calibration. Red, green and blue bars denote x, y and z axes respectively.}
\vspace{-20pt}
\end{center}
\label{fig_jackal}
\end{figure}

On top of the robot base, we designed and manufactured several mounting parts to secure a variety of sensors onboard.
Figure~\ref{fig_jackal} illustrates the sensor setup on the robot.
We discuss the specification and configuration of each sensing modality in detail as follows.

\textbf{Monochrome Camera:} We use the FLIR Blackfly BFS-U3-16S2M-CS camera to capture monochrome images.
It is equipped with a Computar A4Z2812CS-MPIR lens, with an adjustable focal length and aperture.
This monochrome modality can reflect the overall light intensity in the visible spectrum, and is useful for both multimodal perception and vision-based autonomous navigation (e.g., visual odometry~\cite{qin2018vins}).

\textbf{Stereo RGB Camera:} We employ the Stereolabs Zed2i camera to capture color images in visible spectrum. This camera possesses two high-resolution RGB lenses, and depth images can be computed from stereo images in real-time by leveraging the GPU of the Xavier onboard computer. Furthermore, we also include the internal IMU data and the poses computed by the Zed camera driver in our data collection process.

\textbf{Red-Green-NIR Camera:} We use the Mapir Survey3 camera to capture images with three channels: Red, Green, and Near-InfraRed (NIR).
The merit of this multispectral configuration is the capability to compute domain-specific indices directly from a single image. For instance, the Normalized Difference Vegetation Index (NDVI) can be derived from the Red and NIR channels, via the equation $\text{NDVI} = \frac{\text{NIR}-\text{Red}}{\text{NIR}+\text{Red}}$. 
Such indices can enable growers and agronomists to keep track of the health status of plants (e.g.,~\cite{zhao2015ndvi}).

\textbf{Thermal Camera:} We consider the FLIR ADK thermal camera to capture images in the Long-Wave InfraRed (LWIR) range (around 8-14 micrometers). This thermal modality can effectively detect humans and animals, as well as facilitate the understanding of plant health condition (e.g., water stress level, disease prevention, pest control). Furthermore, a thermal modality can also enhance navigation tasks in degenerated environments (e.g.,~\cite{chen2017thermalSLAM}).

A summary of the different cameras used in our robot is provided in Table~\ref{table_camera}.

\begin{table}[h]
\caption{Summary of Camera Specification and Configuration.}
\label{table_camera}
\vspace{-20pt}
\begin{center}
\resizebox{\columnwidth}{!}{
\begin{tabular}{ccccccc}
\toprule
Camera & Modality & Shutter & Rate & Resolution & HFOV & Bit/Channel \\
\midrule
FLIR Blackfly & Monochrome & Global & $10$\;Hz &  $1440 \times 1080$ & 72° & $8 \times 1$\\
FLIR ADK & Thermal & Global  & $10$\;Hz & $640 \times 512$  & 65° & $8 \times 1$ \\
Stereolabs Zed2i & Stereo RGB & Rolling &  $10$\;Hz & $1280 \times 720$ & 102° & $8 \times 3 \times 2$\\
Stereolabs Zed2i & Depth & Rolling & $10$\;Hz &  $1280 \times 720$ & 102° & $32 \times 1$\\
Mapir Survey3 & Red-Green-NIR & Rolling  & $10$\;Hz & $1280 \times 720$  & 85° & $8 \times 3$\\
\bottomrule
\end{tabular}
}
\end{center}
\vspace{-6pt}
{\footnotesize
Note: Although most cameras can support up to $30$\;Hz or $60$\;Hz frame rate, we operate them at $10$\;Hz to match the LiDAR's operating frequency.
}
\end{table}

\textbf{3D LiDAR:} We utilize a Velodyne VLP-16 LiDAR for 3D mapping. It comprises 16 scan lines (constituting 30° vertical FOV) and each line contains 1800 points (constituting 360° horizontal FOV), providing a total of 28800 points per point cloud at a rate of $10$\;Hz. The maximum measurement distance extends to $100$\;m, with an accuracy of $\pm 3$\;cm.

\textbf{IMU:} A MicroStrain 3DM-GX5-AHRS IMU is installed underneath the LiDAR to collect linear acceleration and angular velocity data for inertial navigation. It comes equipped with built-in complementary filter and Extended Kalman Filter (EKF) algorithms that operate ahead of output, at a maximum frequency of $1000$\;Hz. In our configuration, we adjust its rate to $200$\;Hz. The internal reference frames of the accelerometer and gyroscope are treated as identical.

\textbf{Wheel Odometry:} We record the wheel odometry data provided by the Jackal mobile robot software, computed from its wheel encoders at a rate of $50$\;Hz. However, this measurement can drift over time and requires fusion with other sensors for accurate estimation. The reference frame of this sensing modality is located at the bottom center of the robot base. 

\textbf{GPS-RTK:} We utilize the SwiftNav Duro GPS system to function as the positioning ground truth.
By leveraging Real Time Kinematic (RTK) technology, it is possible to acquire centimeter-level positioning accuracy of the robot at a rate of $10$\;Hz. 
The raw measurements, denoted as (latitude, longitude, altitude), are transformed to a path aligned with the origin as per the WGS84 standard, utilizing a Python script.

\section{Sensor Calibration}
The calibration of all sensors is conducted in three steps: 1) calibration of both intrinsic and extrinsic parameters of all cameras, 2) calibration of the extrinsic parameter between the IMU and one selected camera, and 3) calibration of the extrinsic parameter between the LiDAR and one selected camera. We elaborate on each step in the following subsections, and provide a summary discussion at the end of this section.

\begin{figure}[h]
\centering
\includegraphics[width=\linewidth]{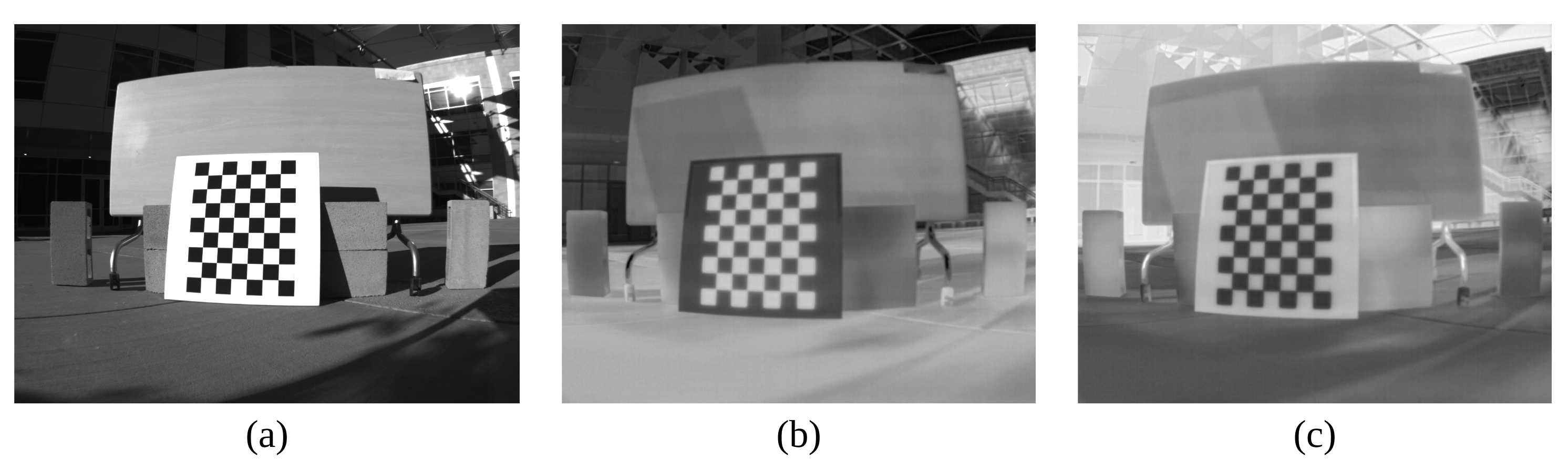}
\caption{Sample images of the calibration target with checkerboard pattern in (a) monochrome modality, (b) thermal modality, and (c) thermal modality with grayscale values inverted. Images are captured under direct sunlight.}
\vspace{-10pt}
\label{fig_calibration_multicam}
\end{figure}

\subsection{Multi-Camera Calibration}
We use the Kalibr~\cite{furgale2013kalibr} toolbox to calibrate four cameras at once. A 24-inch square calibration target with a checkerboard pattern (Figure~\ref{fig_calibration_multicam}(a)) is placed at multiple locations under direct sunlight, to cover most of the field of view of all cameras. Pinhole camera model and radtan (radial-tangential) distortion model are applied in the optimization. Results comprise the intrinsics (including distortion coefficients) of each camera lens, and a chain of transformation matrices from the first to the last camera lens.

One notable challenge we addressed in this process is the recognition of the checkerboard pattern in the thermal modality. In a typical indoor environment, the checkerboard pattern cannot be captured by the thermal camera, since the temperature of the calibration target at each point is almost the same. 
Considering that black areas absorb more heat than white areas, we placed the calibration target outdoor under direct sunlight instead. The resulting pattern being captured by the thermal camera is illustrated in Figure~\ref{fig_calibration_multicam}(b). We then invert its grayscale values to make it consistent with other checkerboard patterns for use in the Kalibr toolbox (see Figure~\ref{fig_calibration_multicam}(c)).

During calibration, the checkerboard pattern in thermal images may become blurry as the heat transfers from black cells to white cells. To alleviate this effect, a foam board was placed in between the poster paper (with the checkerboard pattern printed on it) and the medium density fiberboard (MDF) in the back to reduce heat transmission. Furthermore, the calibration can be conducted in the afternoon when sunlight is not very intense, and a fan can be used to cool down the calibration target when needed.

\subsection{IMU-Camera Calibration}
We use again the Kalibr~\cite{furgale2013kalibr} toolbox to calibrate the extrinsic parameter for the IMU and the FLIR Blackfly monochrome camera.
A 24-inch square calibration target with an aprilgrid pattern is placed on the ground statically. With the pattern being captured by the camera all the time, we first collected a recording of about ten seconds when the robot is static, and then moved the robot (with both sensors rigidly mounted) towards x, y, z, roll, pitch and yaw directions, respectively, each lasting a few seconds. The final transformation matrix is obtained by optimizing the joint observations from the IMU and the camera.

In this process, the intrinsics of both sensors are required. We calibrate the noise density and random walk of the IMU by the Allan variance method beforehand.%
\footnote{\url{https://github.com/ori-drs/allan_variance_ros}}
The intrinsic parameters of the camera are taken from the results of multi-camera calibration.

\subsection{LiDAR-Camera Calibration}
We use the ACFR LiDAR-Camera calibration toolbox~\cite{tsai2021optimising} to estimate the transformation between the LiDAR and the FLIR Blackfly monochrome camera. The 24-inch square calibration target with a checkerboard pattern is reused herein, and is placed at multiple locations within the field of view of both sensors. One additional requirement is that the target has to be placed in a tilted position (e.g., 45-degree) in mid-air, such that the LiDAR point cloud of this target can be segmented out and square edges can be detected. In this way, the calibration program can align the center of the calibration target in both modalities.

\begin{table}[h]
\caption{Summary of Calibrated Extrinsic Parameters.}
\label{table_extrinsics}
\vspace{-20pt}
\begin{center}
\resizebox{\columnwidth}{!}{
\begin{tabular}{lllllllll}
\toprule
Parent Frame & Child Frame & \;$x$[m] & \;$y$[m] & \;$z$[m] & \;$q_x$ & \;$q_y$ & \;$q_z$ & \;$q_w$  \\
\midrule
base\_link & velodyne\_lidar & \;0.0400 & \;0.0000 & \;0.3787 & \;0.0000 & \;0.0000 &  \;0.0000 & \;1.0000 \\
velodyne\_lidar & gps\_rtk & -0.2200 & \;0.0000 & \;0.1530 & \;0.0000 & \;0.0000 &  \;0.0000 & \;1.0000 \\
velodyne\_lidar & flir\_blackfly & \;0.2178 & \;0.0049 & -0.0645 & \;0.5076 &  -0.4989 & \;0.4960 & -0.4974 \\
flir\_blackfly & imu\_link & -0.0061 & \;0.0157 & -0.1895 & \;0.4987 & \;0.5050 & -0.4987 & \;0.4977 \\
flir\_blackfly & zed2i\_rgb\_left & -0.0663 & \;0.0956 & -0.0161 & \;0.0020 & -0.0081 & \;0.0031 & \;1.0000 \\
zed2i\_rgb\_left & zed2i\_rgb\_right &  \;0.1198  & -0.0003 & -0.0046 & \;0.0013 &  \;0.0013 &  \;0.0000 & \;1.0000  \\
zed2i\_rgb\_right & flir\_adk & \;0.0251  & -0.0948 & -0.0203 & \;0.0026 & -0.0032 & \;0.0059 & \;1.0000 \\
flir\_adk & mapir &  -0.1608  & -0.0046 & -0.0138 & \;0.0028 &  \;0.0186 &  -0.0094 & \;0.9998 \\
\bottomrule
\end{tabular}
}
\end{center}
\vspace{-6pt}
{\footnotesize
}
\vspace{-10pt}
\end{table}

\subsection{Summary}
After completing the calibration in these three steps, we can obtain the extrinsic parameters for seven modalities. The remaining two modalities are wheel odometry and GPS-RTK. 
Their extrinsic parameters are represented with respect to the LiDAR sensor, and can be measured and computed from the computer-aided design (CAD) models of the robot base, LiDAR and our mounting plates.

A summary of calibrated extrinsic parameters of all sensors is available in Table~\ref{table_extrinsics}. The seven parameters detailed as ($x$, $y$, $z$, $q_x$, $q_y$, $q_z$, $q_w$) indicate the translation measured in meters and the orientation presented in quaternions. Collectively, these variables define the transformation of a child reference frame with respect to a parent reference frame. The corresponding visualization of all sensor frames using these extrinsic parameters is shown in Figure~\ref{fig_jackal}(b).

\section{Data Collection in Citrus Tree Farms}\label{sec_data_collection}
\begin{figure}[!h]
\centering
\includegraphics[width=\linewidth, trim=30pt 0pt 0pt 0pt, clip]{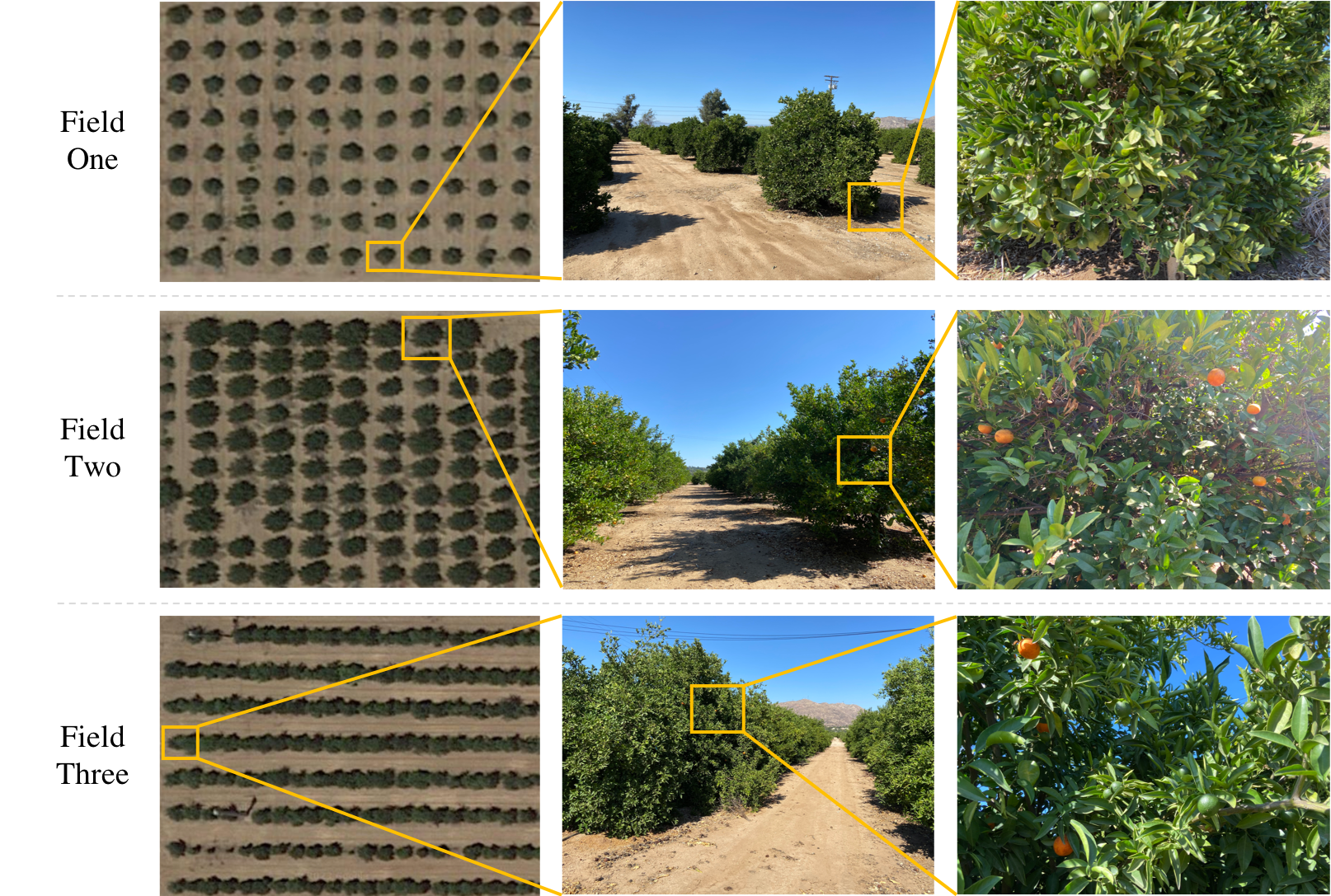}
\caption{The composite satellite image and sample photos of the three fields of citrus trees, covering various planting environments.}
\label{fig_sample_field}
\end{figure}

We collected the CitrusFarm dataset in the Agricultural Experimental Station at the University of California, Riverside in the summer of 2023. 
Three fields of citrus trees were selected in the experiments, owing to their variety in tree species, growth stages and planting patterns. Composite sample images of the three fields are shown in Figure~\ref{fig_sample_field}.

\begin{table}[!h]
\caption{Key Characteristics of Our Developed Dataset.}
\label{table_dataset_ours}
\vspace{-20pt}
\begin{center}
\resizebox{\columnwidth}{!}{
\begin{tabular}{cccccccc}
\toprule
Field & \; \makecell{Planting\\ Pattern} \; & Seq. & \makecell{Distance \\ (m)} & \; \makecell{Duration \\ (min:sec)} \; & \makecell{Time of \\ Day} & Motion & \makecell{Loop or \\Retrace} \\
\midrule
\multirow{3}{*}{One} & \multirow{3}{*}{\makecell{sparse,\\ uniform}} & 01 & 1533.96 & 21:26 & $3$\;pm & lawn-mower & No \\
& & 02 & 524.60 & 07:24 & $5$\;pm & every-fifth-row & No \\
& & 03 & 2035.21 & 28:27 & $7$\;pm  & every-other-row & Yes\\
\midrule
\multirow{2}{*}{Two} & \multirow{2}{*}{\makecell{semi-dense,\\ in-row}} & 04 & 896.15 & 12:49 & $4$\;pm & square-spiral & Yes  \\
& & 05 & 865.33 & 12:39 & $6$\;pm  & lawn-mower & Yes \\
\midrule
\multirow{2}{*}{Three} & \multirow{2}{*}{\makecell{dense,\\ in-row}} & 06 & 356.57 & 05:02 & $4$\;pm & every-other-row & No\\
& & 07 & 1212.74 & 16:53 & $6$\;pm  & lawn-mower & Yes \\
\bottomrule
\end{tabular}
}
\end{center}
\vspace{-10pt}
\end{table}

\begin{figure}[!t]
\centering
\includegraphics[width=0.99\linewidth]{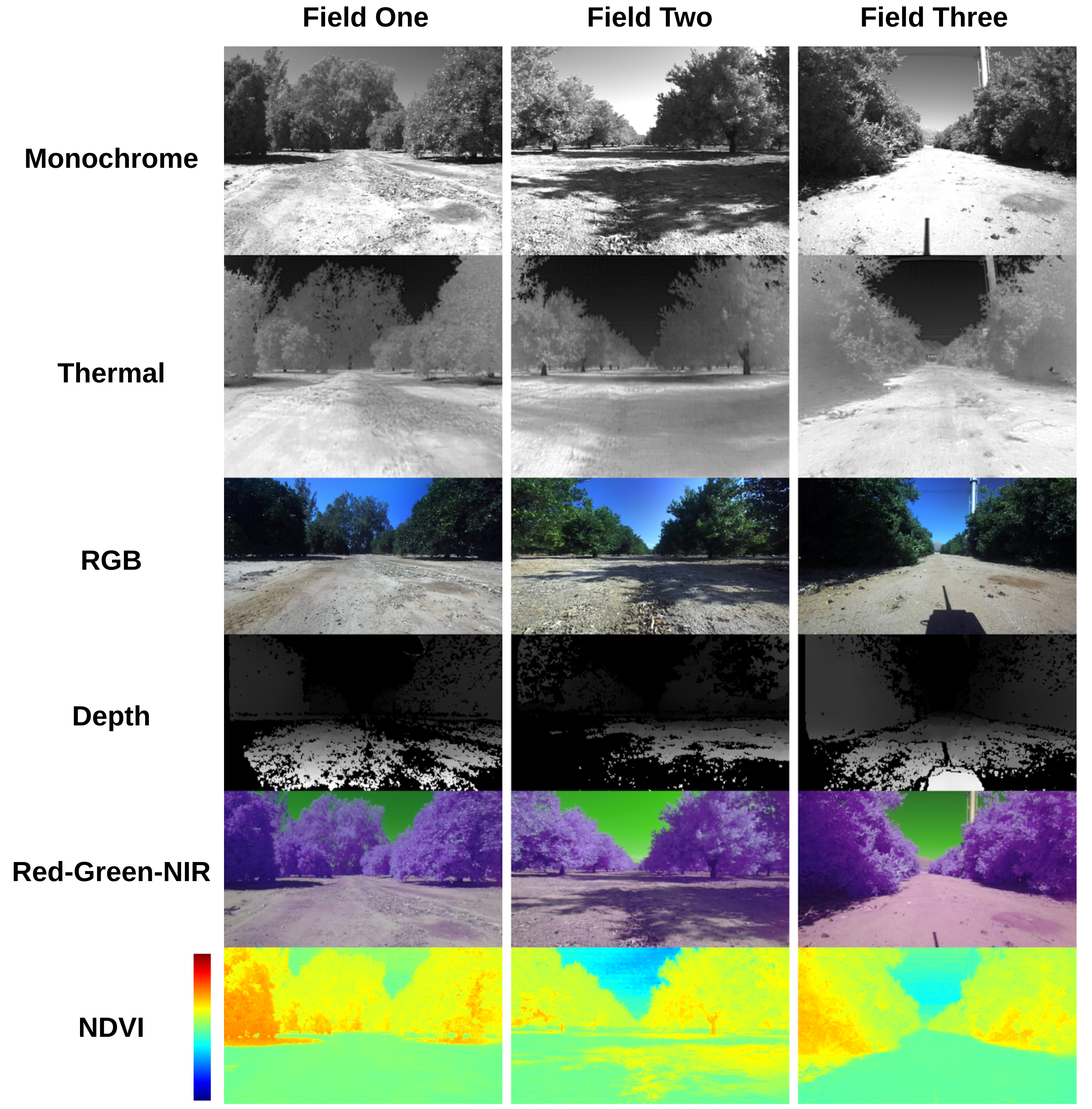}
\vspace{-8pt}
\caption{Sample images collected in the three fields in various modalities, where depth images are computed from stereo RGB images, and NDVI images are computed from the Red and NIR channels of the Red-Green-NIR images.}
\vspace{-14pt}
\label{fig_sample_image}
\end{figure}

During data collection in each field, the robot was remotely piloted and followed by a human operator to travel through various paths.
The GPS-RTK base station, affixed to a tripod, was positioned in the center of the field to maintain proximity to the robot.
By leveraging high-gain antennas for both the base station and the rover receiver on the robot, the reliable communication range can be increased to $150$\;m, with trees interposed. Notably, no instances of signal loss were observed throughout the experiments.

\begin{figure}[!t]
\centering
\includegraphics[width=\linewidth]{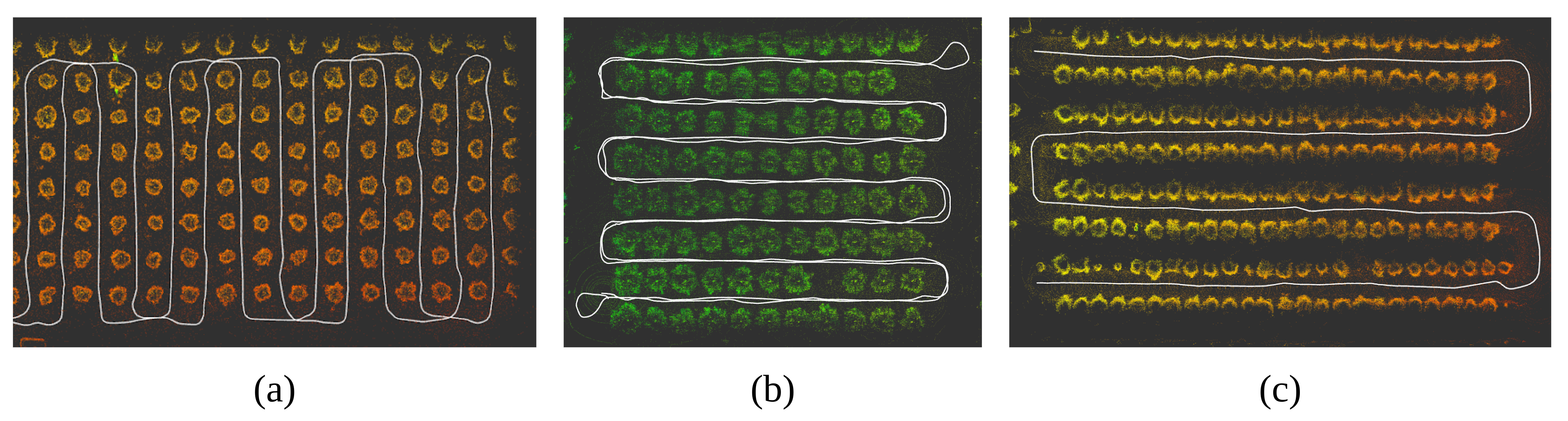}
\vspace{-20pt}
\caption{Sample point cloud maps reconstructed from 3D LiDAR data, superimposed with GPS-RTK trajectories, for (a) sequence 03, (b) sequence 05, and (c) sequence 06 (sequence details are listed in Table~\ref{table_dataset_ours}).}
\label{fig_sample_lidar_gps}
\vspace{-14pt}
\end{figure}

In total, we collected seven sequences in the three fields. These sequences span a total operation time of $1.7$\;hours, cover a total distance of $7.5$\;km, and constitute a total of $1.3$\;TB of data. Key characteristics of the dataset are summarized in Table~\ref{table_dataset_ours}.
The dataset offers a total of nine sensing modalities. Figure~\ref{fig_sample_image} depicts sample images from five camera modalities, and one instance of a domain-specific index computed from multispectral images (i.e. NDVI). Figure~\ref{fig_sample_lidar_gps} provides sample trajectories of selected sequences.

The primary data format we used in data collection is ROS bags. To simplify data storage and transfer, we split recorded data into blocks of 4GB, and categorized them based on their respective modalities. 
This categorization allows users to selectively download the subsets that are of their interest.
To utilize the dataset, users need to only place the downloaded subsets in a common folder and initiate data playback; ROS will then automatically arrange the data across all bags and sequence the playback according to timestamps. 
Additional details on how to utilize the dataset, along with some highlighted use cases, are provided in the open-source repository for this work.

Further, to accommodate diverse use cases, we also provide Python scripts to extract data from ROS bags into files, organized in different levels of folders. These comprise images of five modalities, point clouds obtained from 3D LiDAR, CSV files for IMU readings, wheel odometry readings, and ground-truth paths from the GPS-RTK data.
Moreover, raw data collected during the calibration process and the computed calibration parameters are included in the dataset.

\vspace{-2pt}
\section{Conclusion}
In this work we presented the CitrusFarm dataset, a comprehensive multi-sensor data collection in the citrus tree farms. 
The dataset comprises seven sequences collected in three fields of citrus trees, featuring various tree species at different growth stages, distinctive planting patterns, as well as varying daylight conditions. 
Furthermore, it provides a total of nine sensing modalities: monochrome, stereo RGB, depth, near-infrared and thermal images, as well as wheel odometry, LiDAR, IMU and GPS-RTK data.

We anticipate that this dataset can facilitate the development of autonomous robot systems operating in agricultural tree environments, primarily for localization, mapping and navigation tasks.
The rich sensing modalities provided in this dataset can promote a wide range of research in robotics and computer vision, such as sensor fusion and multimodal machine learning.

Future works based on this dataset can include the development of a multimodal mapping framework for crop monitoring, sensor fusion for localization in unstructured agricultural tree environments, and the integration of localization and mapping algorithms into fully automated agricultural operations.

\vspace{-2pt}
\section*{Acknowledgments}
\vspace{-2pt}
We gratefully acknowledge the support of NSF \# CMMI-2046270, USDA-NIFA \# 2021-67022-33453, ONR \# N00014-18-1-2252 and Univ. of California UC-MRPI \# M21PR3417. 
Any opinions, findings, and conclusions or recommendations expressed in this material are those of the authors and do not necessarily reflect the views of the funding agencies.

\vspace{-4pt}
%

\bibliographystyle{splncs04}
\bibliography{refs}

\end{document}